\PassOptionsToPackage{table}{xcolor}
\documentclass[runningheads]{llncs}

\usepackage[preprint]{colm}

\setcitestyle{authoryear}

\usepackage{graphicx}
\usepackage{booktabs}
\usepackage[accsupp]{axessibility}
\usepackage{xltabular} 
\usepackage{microtype}
\usepackage{tabularx}
\usepackage{amssymb}
\usepackage{paralist}
\usepackage{todonotes}
\usepackage{breakurl}
\usepackage{subfig}
\usepackage{latexsym}
\usepackage{amsmath}
\usepackage{float}
\usepackage{footnote}
\usepackage{enumitem}
\usepackage{bm}
\usepackage{arydshln}
\usepackage{multicol}
\usepackage{multirow}
\usepackage{colortbl}
\usepackage{bbding}
\usepackage{makecell}
\usepackage{mathtools}
\usepackage{imakeidx}
\usepackage{longtable}
\usepackage{tabularx}
\usepackage{wrapfig}
\usepackage{rotating}
\usepackage[edges]{forest}
\usepackage[normalem]{ulem}
\usepackage{CJKutf8}
\usepackage{awesomebox}
\usepackage[most]{tcolorbox}
\usepackage{graphicx}
\usepackage{booktabs}
\usepackage[table]{xcolor}
\usepackage{adjustbox}

\RequirePackage{xspace}
\makeatletter
\DeclareRobustCommand\onedot{\futurelet\@let@token\@onedot}
\def\@onedot{\ifx\@let@token.\else.\null\fi\xspace}

\makeatother

\usepackage{pifont}
\usepackage{lscape}
\usepackage{algorithm}
\usepackage{algpseudocode}

\usepackage{fancyhdr}
\renewcommand{\headrulewidth}{1pt}
\usepackage{fancyvrb}
\usepackage{fvextra}
\usepackage{amsfonts}
\usepackage{wrapfig}

\usepackage{hyperref}
\usepackage{orcidlink}

\definecolor{mydarkblue}{rgb}{0,0.08,0.45}
\definecolor{wkblue}{rgb}{0.2, 0.3, 0.6}
\definecolor{meta-color}{rgb}{0.5, 0.5, 0.5}
\definecolor{darkblue}{rgb}{0, 0, 0.5}
\definecolor{geovistagray}{gray}{0.95}
\definecolor{myblue}{rgb}{0.9, 0.1, 0.94}
\definecolor{mygreen}{rgb}{0.64, 0.56, 0.88}
\definecolor{myyellow}{rgb}{0.68, 0.6, 0.1}
\definecolor{fancygreen}{rgb}{0.33, 0.68, 0.20}
\definecolor{salmon}{rgb}{0.94, 0.52, 0.49}
\definecolor{tablegreen}{rgb}{0.82, 0.94, 0.75}
\definecolor{tableblue}{rgb}{0.81, 0.90, 0.94}
\definecolor{tablered}{rgb}{0.97, 0.85, 0.85}
\definecolor{tableorange}{rgb}{0.96, 0.85, 0.81}
\definecolor{bestcolor}{RGB}{210, 222, 239}
\definecolor{secondcolor}{RGB}{234, 239, 247}
\definecolor{thirdcolor}{RGB}{193, 214, 229}
\definecolor{line-blue}{RGB}{243, 248, 252}
\definecolor{line-green}{RGB}{200,242,200}
\definecolor{line-red}{RGB}{255,215,215}
\definecolor{line-gray}{RGB}{242, 242, 242}
\definecolor{sensepurple}{HTML}{5D2DD6}

\newcommand{\paragrapha}[1]{\par\addvspace{0.45em}\noindent\textbf{#1}}

\begin{document}

\title{ViQ: Text-Aligned Visual Quantized Representations at Any Resolution}

\titlerunning{ViQ}

\author{\textbf{Xumin Yu$^{1,*}$, Zuyan Liu$^{1,2,*}$, Zhenyu Yang$^{1,4,*}$, Yuhao Dong$^{3}$,} \\
\textbf{Shengsheng Qian$^{4}$, Jiwen Lu$^{2,\dagger}$, Han Hu$^{1}$, Yongming Rao$^{1,\dagger}$}\\[0.5em]
$^1$ Tencent HY Vision Team \quad $^2$ Tsinghua University \\[0.2em]
$^3$ Nanyang Technological University \quad $^4$ Institute of Automation, CAS}

\authorrunning{Yu et al.}

\maketitle

\makeatletter
\def\@makefnmark{}
\makeatother
\footnotetext{$^{*}$ Equal contribution. $^{\dagger}$ Corresponding author.}

\begin{flushleft}
\small
\textbf{GitHub:} \href{https://github.com/yuxumin/ViQ}{\nolinkurl{https://github.com/yuxumin/ViQ}} \\
\textbf{HuggingFace:} \href{https://huggingface.co/XuminYu/ViQ_weights}{\nolinkurl{https://huggingface.co/XuminYu/ViQ_weights}}
\end{flushleft}
\vspace{0.3em}

\begin{abstract}
A unified representation for text and vision is a natural pursuit, as it enables simpler multimodal modeling and more efficient training. However, representing images as discrete signals in the same way as text inevitably introduces severe information loss. Existing work struggles to balance low-level details and high-level semantics in discrete representations: reconstruction-oriented representations often lack semantic information, whereas semantically stronger features typically suffer from severe loss of detail. We present \textbf{ViQ}, a \textbf{Vi}sual \textbf{Q}uantized Representations framework, which is designed to balance semantics and details in discrete representations while supporting inputs at native resolutions, thereby enabling it to serve as a unified and general discrete representation for arbitrary visual inputs. Our approach structures quantization learning into two stages: text-aligned pre-training and feature discretization. With text-aligned pre-training, we enhance the visual encoder semantic-rich supervision from the pretrained language model and enable it to process native-resolution visual inputs. During discretization, we propose a proximal representation learning strategy to progressively compact the feature space, along with a position-aware head-wise quantization mechanism that enables flexible processing of arbitrary resolutions. Extensive experiments on multimodal tasks demonstrate that ViQ achieves competitive performance compared to state-of-the-art multimodal vision encoders with continuous and high-dimensional visual features, while maintaining high precision in low-level reconstruction. We also show that multimodal training with visual quantized representations largely improves efficiency, yielding up to 20\%-70\% acceleration with different base LLMs and training recipes.
\end{abstract}

\section{Introduction}
\label{sec:intro}

The rapid development of multimodal large language models (MLLMs)~\cite{qwen2vl,chen2024internvl,zhu2025internvl3} has created a growing demand for high-quality, unified visual representations. A key challenge in this field lies in aligning visual signals into a form that can be seamlessly interpreted by language models. Early and dominant approaches have primarily relied on continuous visual encoders~\cite{zhai2023siglip, radford2021clip, fini2025multimodal}, pre-trained through contrastive vision-language learning, or specialized encoders fine-tuned for multimodal tasks~\cite{chen2412expanding,qwen2vl,liu2024oryx}. Although this paradigm has achieved considerable success, it introduces a fundamental representational discrepancy: the continuous nature of visual features is intrinsically mismatched with the discrete token-based modeling of text. Furthermore, the computational process of extracting high-dimensional visual features places significant hardware strain during model training.

\begin{figure}[t]
\centering
\includegraphics[width=1.0\linewidth]{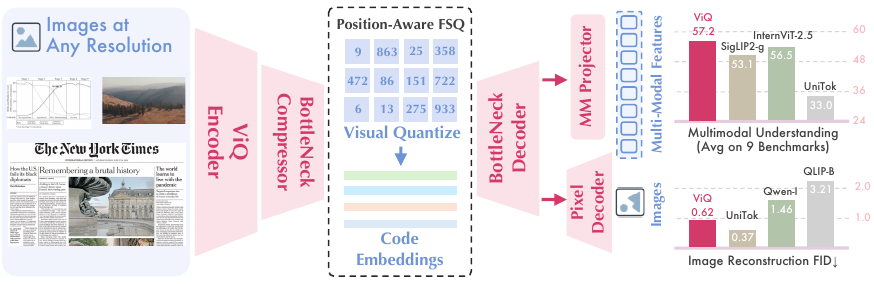}\vspace{-5pt}
\caption{\textbf{ViQ delivers high-quality multimodal quantized representations with both high-level semantics and low-level details.} The quantized visual codes in ViQ support high-level multimodal understanding and low-level image reconstruction, with state-of-the-art performance compared with continuous visual encoders.}
\label{fig:teaser}
\end{figure}

Inspired by the success of vector quantization~\cite{gersho2012vector,mentzerfinite,yulanguage} in visual representation learning, a growing body of research has begun exploring discrete visual representations in multimodal domains~\cite{ma2025unitok,zhao2025qlip,meituanlongcatteam2026longcatnextlexicalizingmodalitiesdiscrete}. These approaches typically quantize visual inputs into a finite set of discrete tokens, analogous to words in a textual vocabulary. However, a fundamental challenge remains in balancing low-level visual details with high-level semantic information. Reconstruction-focused autoencoders often fail to preserve semantic structure in their latent features, while semantically rich visual encoders tend to incur significant information loss during quantization. This creates a critical gap in multimodal representations, which require both fine-grained visual details and rich semantic content to handle complex tasks effectively. As a result, the application of quantized visual encoders has been largely limited, and they have yet to match the high fidelity and robustness of continuous encoders in demanding real-world scenarios.

To address this issue, we propose ViQ (Visual Quantized Representations), a novel approach that elevates the performance of discrete visual representations in multimodal tasks to a level comparable with widely-used continuous features, while supporting native-resolution visual inputs. Our method structures quantization learning into two phases: text-aligned pre-training and visual quantization learning. In the initial stage, the ViQ model learns semantically rich alignments through supervision from vision-language pairs. To mitigate information loss during quantization, we introduce a proximal representation learning strategy that constrains the latent visual space, followed by a quantization mechanism enhanced with position encoding and expanded visual features. This design supports quantization at arbitrary resolutions and improves the representational capacity of visual codes. Our work not only strongly enhances the efficiency of multimodal encoders in training process but also unifies visual and linguistic representations into a cohesive discrete framework.

We evaluate the ViQ model against state-of-the-art continuous multimodal visual encoders and quantized multimodal semantic encoders across a range of experiments. Under consistent fine-tuning data and training protocols, ViQ not only outperforms existing quantized models by a large margin but also achieves competitive results compared to well-established continuous encoders such as InternViT~\cite{zhu2025internvl3}, AIMv2~\cite{fini2025multimodal}, and SigLIP2~\cite{tschannen2025siglip}. On the aggregated score over nine benchmarks—spanning visual question answering, world knowledge, and document and chart recognition—ViQ attains an average of 57.2 with Qwen2.5-1.5B as the backbone LLM and 63.9 with Qwen2.5-7B, surpassing the previous state-of-the-art scores of 57.0 and 63.8, respectively, under 6B number of visual encoder parameters. The quantized representation also brings substantial efficiency gains in real-world multimodal training. Compared to conventional training strategies, using the ViQ visual encoder yields speedups of 20\% to 70\% in training recipes varying in sequence lengths. Furthermore, ViQ preserves rich low-level visual details: when fine-tuned with an image decoder, it achieves a PSNR of 22.73 and an rFID score of 0.62, ranking first among mainstream discrete visual autoencoders. This strong performance in both understanding and reconstruction tasks underscores the effectiveness and unity of our discrete representation.

\section{Related Works}
\label{sec:related_works}

\paragrapha{Visual Representations for MLLMs. }The visual encoder serves as a critical bridge between vision and language modalities in multimodal learning. Conventional multimodal language models typically employ CLIP-style models—such as CLIP~\cite{radford2021clip}, SigLIP~\cite{zhai2023siglip}, SigLIP2~\cite{tschannen2025siglip}, AIM~\cite{fini2025multimodal}, and DFN~\cite{fang2023dfn}—as visual encoders. However, such models often rely on fixed input resolutions, which constrains their flexibility in multimodal tasks, and their contrastively learned representations may not align optimally with downstream multimodal objectives. To address these limitations, several specialized MLLM visual encoders have been developed and trained end-to-end on multimodal tasks. These include open-source models like InternViT~\cite{chen2024internvl,zhu2025internvl3}, OryxViT~\cite{liu2024oryx}, and SAILViT~\cite{yin2025sail}, as well as proprietary encoders integrated into large MLLM systems such as Qwen-VL~\cite{Qwen2.5-VL}, Kimi-VL~\cite{team2025kimi}, and Seed-VL~\cite{guo2025seed1}. More recently, researchers have begun exploring unified discrete representations for vision and language. Quantized multimodal encoders such as QLIP~\cite{zhao2025qlip} and UniTok~\cite{ma2025unitok} apply quantization to image inputs. Nevertheless, these quantized visual encoders still exhibit a large performance gap compared to continuous models, particularly in tasks requiring textual understanding or fine-grained visual details. This gap can be attributed to the high compression ratio of discrete codes and the high sensitivity of multimodal models to detailed visual information.

\paragrapha{Quantized Visual Encoders. } Image tokenization is pivotal in bridging raw pixels with compact latent representations for visual generation. Among existing approaches, vector-quantized (VQ)~\cite{gersho2012vector} tokenizers have gained prominence due to their discrete latent space, which is well-suited for visual generation. The seminal VQ-VAE~\cite{van2017neural} introduced a learnable codebook to discretize continuous features via nearest-neighbor assignment. Subsequent methods (such as FSQ~\cite{mentzerfinite}, RPQ~\cite{chiu2022self}, BSQ~\cite{zhaoimage}, and LFQ~\cite{yulanguage}) explored structurally constrained codebooks with predefined geometries, often decoupled from end-to-end training. Enhancements like VQ-GAN~\cite{esser2021taming} further improved visual quality by incorporating adversarial and perceptual losses. While these tokenizers excel at learning compact, generative-friendly representations, they often struggle to preserve fine-grained visual details crucial for dense prediction and high-fidelity understanding tasks. To address this, we propose ViQ, a novel framework designed to minimize information loss and align discrete visual tokens with semantic and textual contexts.

\section{ViQ: Visual Quantized Representations}
\label{sec:method}

\begin{figure*}[t]
\centering
\includegraphics[width=\linewidth]{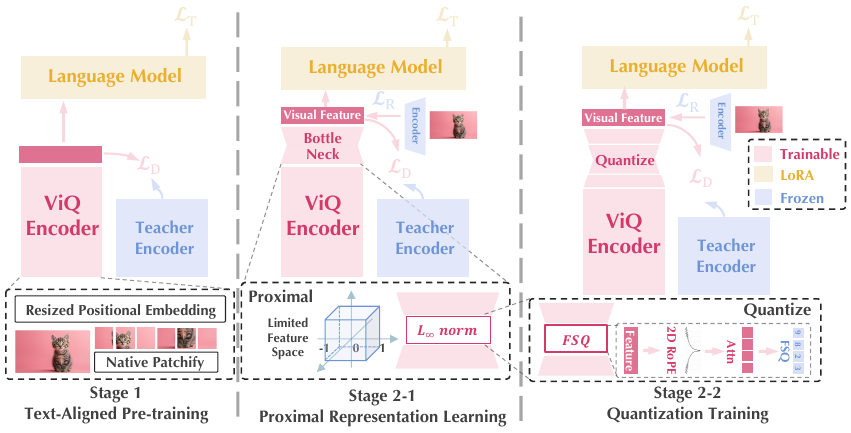}
\caption{\textbf{Approach of ViQ Representation Learning.} Stage 1 enables multimodal alignment with language supervision, while Stage 2 compresses the high-dimensional visual features into discrete codes in a progressive learning manner. }
\label{fig:approach}
\end{figure*}

The visual quantization training for ViQ follows a two-stage process, as depicted in Fig.~\ref{fig:approach}. In Stage 1, text-aligned pre-training is performed on continuous features to enhance multimodal alignment. Subsequently, Stage 2 applies quantization in a progressive manner.

\subsection{Text-Aligned Pre-Training at Any Resolution}
\label{sec:methods_1}

The first stage of ViQ training aims to create a visual encoder that functions in a multi-modal manner. This is accomplished by leveraging text-aligned visual pre-training to align the vision and language embeddings within ViQ, thereby following an optimization strategy similar to conventional multi-modal training.

\paragrapha{Any Resolution Adaptation. }While Language-Image Pre-training models such as CLIP~\cite{radford2021clip} and SigLIP~\cite{zhai2023siglip} offer a strong initialization for building multi-modal visual encoders, most existing models are constrained by a fixed input size. In multi-modal learning, native image resolution often provides a more natural and efficient form of visual representation. To overcome the limitations of fixed-scale pre-training, we follow the approach of NaViT~\cite{dehghani2024navit} and replace the original positional embedding layer with one that supports an adequately dimensioned size. This allows the model to resize positional parameters when processing images at arbitrary resolutions dynamically. We adopt the techniques introduced in OryxViT~\cite{liu2024oryx} to maintain computational efficiency with variable-length visual tokens.

\paragrapha{Text-Guided Multi-Modal Pre-training. }We optimize the continuous feature space of ViQ to be more compatible with multi-modal learning during our text-aligned pre-training stage. Specifically, given a data triplet $[I, T, A]$ consisting of an image $I$, a text query $T$, and an answer $A$, we employ a temporary language model to compute the text supervision loss $\mathcal{L}_{\text{text}}$, which is defined as:
\begin{equation}
    \mathcal{L}_{\text{text}}=\text{Cross Entropy}[\ \text{LLM}(\text{ViQ}(I),\ T),\ A]
\end{equation}

\paragrapha{Self Distillation. }During the multi-modal pre-training stage, we apply self-distillation to regularize the ViQ model, preventing it from overfitting to the multi-modal data and disregarding the knowledge acquired from large-scale language-image pre-training, which is crucial for maintaining the generalizability and foundational capabilities. We use the original fixed-resolution model as the teacher to supervise the cosine similarity of the semantic token (i.e., the class token in mainstream architectures), ensuring that the representations produced at any resolution preserve the original high-level semantic information:
\begin{equation}
    \mathcal{L}_{\text{distill}} = 1 - \cos\left(\mathbf{z}_s^{\text{student}}, \mathbf{z}_s^{\text{teacher}}\right)
\end{equation}
where \(\mathbf{z}_s^{\text{student}}\) and \(\mathbf{z}_s^{\text{teacher}}\) denote the semantic tokens from the student (any-resolution) and teacher (fixed-resolution) models, respectively.

\paragrapha{Training Recipe for Native Resolution. }We employ a progressive training strategy that transitions from fixed-resolution to any-resolution inputs. Specifically, the training begins with low-resolution images at their native aspect ratios, and the total number of pixels is gradually increased throughout the training process until native resolution is attained. During training, multi-modal data are packed to fit the maximum token length for higher pre-training efficiency. The image complexity, question-answering difficulty, and overall sequence length are progressively raised to steadily adapt the model to multi-modal tasks. Further implementation details are provided in the appendix. We combine the text loss \(\mathcal{L}_{\text{text}}\) and the distillation loss \(\mathcal{L}_{\text{distill}}\) for joint optimization.

\subsection{Visual Quantized Representation Learning}
\label{sec:methods_2}

The second stage of ViQ training involves quantizing the continuous features into discrete codes. A core challenge in this process is to fully optimize the discrete representation space so as to preserve fine-grained visual information. Through our approach, we demonstrate that the resulting ViQ representations are effective in supporting high-quality visual understanding in multimodal contexts.

\paragrapha{Proximal Representation. }The primary objective of quantization is to map continuous visual features from a high-dimensional space $\mathbb{R}^{C}$ into a constrained discrete space defined by a codebook $\mathcal{Z}$. However, we observe that directly quantizing high-dimensional visual features results in significant precision loss. To mitigate this, our proposed ViQ method progressively reduces the complexity of the latent space using a proximal representation. Given a high-dimensional visual feature $f \in \mathbb{R}^C$, we first apply a bottleneck layer to compress it into an intermediate-dimensional feature $f_1 \in \mathbb{R}^D$. We then further constrain the feature space complexity via a regularization function, referred to as a proximal representation. In our implementation, we apply the $L_\infty$-norm to the compressed latent feature, projecting all features onto a hypercube surface such that $\lVert f_1 \rVert_\infty = 1$. This step progressively reduces the feature space complexity prior to discrete quantization. The proximal representation helps regulate the distance between features and quantization anchors, thereby reducing information loss during quantization. The bottleneck compression operations can be formulated as:
\begin{equation}
\begin{aligned}
    f_1=L_{\infty}(\text{BN}(f)), \hat{f}=\text{BN}'(f_1)
\end{aligned}
\end{equation}
where $\text{BN}$ denotes the bottleneck fully connected layer for dimension compression and $\text{BN}'$ denotes the inverted bottleneck layer respectively.

\paragrapha{Multi-Head Finite Scalar Quantization. }After learning an initial continuous proximal representation within a constrained latent space, we replace the normalization function with a bottleneck downsampling layer that projects the latent features into a lower-dimensional space \( f_2 \in \mathbb{R}^d \), where \( d \ll D \ll C \). The final dimension \( d \) is kept sufficiently small to facilitate effective quantization. We employ Finite Scalar Quantization (FSQ)~\cite{mentzerfinite} to discretize the compact continuous features \( f_2 \), as this method offers stable training behavior without requiring additional optimization during quantization. The FSQ process can be formulated as:
\begin{equation}
z = \text{round}(\mathcal{Q}(f_2))
\end{equation}
where $\mathcal{Q}$ indicates the Finite Scalar Quantization function.

To enhance the representational capacity of the quantized codes, we utilize a multi-head attention mechanism that expands each visual patch into \( 2 \times 2 \) visual codes. Specifically, the latent feature \( f_2 \in \mathbb{R}^{B \times N \times d} \) is up-projected to \( \mathbb{R}^{B \times 4N \times d} \), where \( N \) denotes the number of original image patches. Here the up-projection is a linear layer \( \mathbb{R}^{d}\!\to\!\mathbb{R}^{4d} \) that lifts each patch and reshapes it into \( 4 \) sub-tokens. The multi-head self-attention is applied only \emph{within} the \( 4 \) sub-tokens of the same patch, so that different patches never interact; quantization is then performed element-wise on these tokens. Finally, the \( 4 \) quantized sub-tokens of each patch are concatenated and passed through a linear projection \( \mathbb{R}^{4d}\!\to\!\mathbb{R}^{d} \) to restore one token per patch and maintain the original downsampling rate. It is important to note that the expanded image patches are processed independently; as a result, the quantized representations in ViQ remain independent, making them more suitable for representation learning and downstream tasks.

\paragrapha{Rotary Position Embedding. }For quantized representations at arbitrary resolutions, we incorporate a 2D rotary position embedding (RoPE)~\cite{su2024roformer} to explicitly encode spatial resolution information. The RoPE layer is inserted prior to the quantization step. Specifically, 2D RoPE extends the 1D RoPE formulation used in large language models by embedding both height and width information of the visual content. Given a token feature \( f_m \in \mathbb{R}^d \) at position \( (h, w) \) in the 2D feature map, the rotary encoding is applied as:
\begin{equation}
\tilde{f}_m = f_m \odot e^{i (h \theta_h + w \theta_w)}
\end{equation}
where \( \theta_h \) and \( \theta_w \) are frequency parameters for the height and width dimensions, respectively, and \( \odot \) denotes element-wise multiplication with complex exponentials. This formulation ensures that relative spatial relationships are preserved across varying resolutions.

\paragrapha{Multi-Stage Training with Low-Level Supervision. }Building upon the progressive quantization framework, we learn the compressed feature representation in a multi-stage manner. First, starting from the proximal representation in the constrained space, we introduce a bottleneck layer with a regularization function between the continuous input features and the final output. We retain both the text loss \(\mathcal{L}_{\text{text}}\) and the self-distillation loss \(\mathcal{L}_{\text{distill}}\) during this phase, keeping all other settings consistent with Stage 1 training. To enhance low-level detail preservation and improve latent feature quality, we incorporate a pre-trained visual autoencoder that encodes the original input images into the low-level latent features. This encoding is supervised using a VAE-style latent loss along with a negative log-likelihood (NLL) term, expressed as:
\begin{equation}
\mathcal{L}_{\text{recon}} = \text{NLL}(\hat{f}, \text{Encoder}(x))
\end{equation}
where \(x\) denotes the input image and $\hat{f}$ denotes the recovered features after compression or quantization. Here the NLL term is computed under a Gaussian likelihood with fixed (unit) variance over the target VAE latent $\text{Encoder}(x)$, so that the loss reduces, up to a constant, to a mean-squared error between $\hat{f}$ and $\text{Encoder}(x)$, i.e., $\mathcal{L}_{\text{recon}} = \tfrac{1}{2}\lVert \hat{f} - \text{Encoder}(x)\rVert_2^2 + \text{const}$. This makes the objective a simple and stable regression on the pre-trained VAE latent space rather than a pixel-level reconstruction.

Once the bottleneck compression is adequately learned, we replace the proximal regularization function with a quantization module in the subsequent quantization training stage. Since our vector quantization (VQ) implementation is optimization-free, no additional codebook supervision is required at this stage. The overall training objective combines the three losses as follows:
\begin{equation}
\mathcal{L}_{\text{total}} = \lambda_{\text{text}}\mathcal{L}_{\text{text}} + \lambda_{\text{distill}}\mathcal{L}_{\text{distill}} + \lambda_{\text{recon}}\mathcal{L}_{\text{recon}}
\end{equation}

\section{Experiments}
\label{sec:experiments}

In our experiments, we comprehensively evaluate the effectiveness of the ViQ model. We first describe the implementation details of ViQ's key components. We then perform fair comparisons on multimodal tasks under consistent fine-tuning data and training protocols, benchmarking against leading multimodal and quantized encoders. To highlight practical benefits, we assess efficiency in real-world training and visual storage scenarios. We further examine the reconstruction quality to analyze low-level feature preservation, and provide ablation studies to validate design choices.

\subsection{Implementation Details}

We instantiate our ViQ model based on the pretrained SigLIP2-g~\cite{tschannen2025siglip} visual encoder, whose output feature dimension is $C=1536$. To meet quantization requirements, we reduce the dimensionality through bottleneck layers to $D=128$ and subsequently to $d=6$. For quantization, we employ the FSQ method~\cite{mentzerfinite} with levels $\mathcal{L}=[8,8,8,5,5,5]$, yielding a codebook size of 64,000. Multimodal features from ViQ are projected via a simple MLP layer. During training, we use Qwen2.5-VL-0.5B as the language model for text supervision and the pretrained Qwen-Image~\cite{wu2025qwenimage} encoder for low-level visual supervision, keeping its parameters fixed. Stage 1 pretraining is conducted using 128 NVIDIA A100 GPUs, while Stage 2 quantization training uses 256 NVIDIA A100 GPUs. Further details regarding training stages and datasets are provided in the appendix.

\begin{table*}[t]
  \centering
  \caption{\textbf{Comparison results on multi-modal understanding tasks.} We conduct experiments with various visual encoder counterparts on general multi-modal benchmarks. All the methods are trained with the same data collection.}\vspace{10pt}
  \begin{adjustbox}{width=\linewidth}
    \begin{tabular}{lccccccccccccc}
    \toprule
    \multicolumn{4}{c}{Base LLM}        & \multicolumn{1}{c}{\multirow{2}[4]{*}{\rotatebox{30}{MMStar}}} & \multicolumn{1}{c}{\multirow{2}[4]{*}{\rotatebox{30}{MMMU}}} & \multicolumn{1}{c}{\multirow{2}[4]{*}{\rotatebox{30}{SimpleVQA}}} & \multicolumn{1}{c}{\multirow{2}[4]{*}{\rotatebox{30}{InfoVQA}}} & \multicolumn{1}{c}{\multirow{2}[4]{*}{\rotatebox{30}{TextVQA}}} & \multicolumn{1}{c}{\multirow{2}[4]{*}{\rotatebox{30}{DocVQA}}} & \multicolumn{1}{c}{\multirow{2}[4]{*}{\rotatebox{30}{OCRBench}}} & \multicolumn{1}{c}{\multirow{2}[4]{*}{\rotatebox{30}{AI2D}}} & \multicolumn{1}{c}{\multirow{2}[4]{*}{\rotatebox{30}{ChartQA}}} & \multicolumn{1}{c}{\multirow{2}[4]{*}{\rotatebox{30}{Avg.}}} \\
\cmidrule{1-4}    \multicolumn{1}{c}{Visual Encoder} & Size  & AnyRes & Discrete &       &       &       &       &       &       &       &       &       &       \\
    \midrule
    \multicolumn{4}{c}{\textit{Qwen2.5 - 1.5B}} \\
    \midrule
    OAI-CLIP-L~\cite{radford2021clip} &   0.3B    & \ding{55}  &   \ding{55}   &44.9 &	40.7 	&21.3 &	24.2 	&58.9 &	52.9 &	460.0 &	68.1 	&55.0 &	45.8 \\
    SigLIP2-g~\cite{tschannen2025siglip} & 1.1B    & \ding{55}   &\ding{55}  &48.1 	 &42.4  &	25.6  &	28.2 	 &73.1  &	67.8  &	590.0  &	71.5  &	62.0 	 &53.1  \\
    DinoV2-g~\cite{oquab2023dinov2} &   1.1B    & \ding{55}     &   \ding{55}     &     47.1 	&41.8 	&24.0 	&31.7 	&72.1 &	76.9 	&619.0 	&68.8 	&61.8 	&54.0 \\
    \midrule
    OryxViT~\cite{liu2024oryx}  & 0.4B  &  \ding{51}      &   \ding{55}     &    46.4   &  42.1     &   23.2    &   31.8    &   71.8    &   73.5    &   622.0    &    68.2   &    62.1   &   53.4  \\
    AIMv2-H~\cite{fini2025multimodal} &   0.7B    & \ding{55}    &   \ding{55}     & 48.5 &	41.8 &	23.5 	&31.9 	&71.6 	&73.7 &	623.0 &	69.8 &	62.5 &	53.9   \\
    InternViT-2.5~\cite{chen2412expanding} & 0.3B  & \ding{55}     &   \ding{55}     & 47.9&  	40.3 	& 23.6 	& 35.5 & 	73.0 & 	81.7 & 	681.0 & 	69.6 & 	\textbf{69.2} & 	56.5  \\
    InternViT-2.5-6B~\cite{chen2412expanding} & 6.0B & \ding{55}     &   \ding{55}     & \textbf{48.5}&  	42.1 & 	23.7 	& 35.2 & 	\textbf{75.5} & 	80.1 & \textbf{690.0} & 	\textbf{70.7} & 	67.8 & 	57.0  \\
    \midrule
    QLIP~\cite{zhao2025qlip}  &   0.3B    & \ding{55}    &    \ding{51}  &39.9 &	36.9 &	13.7 &	14.8 &	45.1 &	12.2 &	290.0 &	61.9 	&14.1 &	29.7 \\
    UniTok~\cite{ma2025unitok} &   0.3B    &  \ding{55}      &   \ding{51}  &  41.0 &	36.1 &	15.5 	&15.9 &	39.7 &	11.6 	&323.0 &	61.2 &	43.8 &33.0  \\
    \midrule
    \rowcolor{red!10} ViQ & 1.3B  &   \ding{51}     &   \ding{51}     & 47.8  & \textbf{42.6}  & \textbf{26.0}  & \textbf{41.6}  & 74.3  & \textbf{84.2}  & 636.0  & 69.7  & 65.2  & \textbf{57.2}\\
    \midrule
    \multicolumn{4}{c}{\textit{Qwen2.5 - 7B}} \\
    \midrule
    OAI-CLIP-L~\cite{radford2021clip} &    0.3B   &  \ding{55}      &     \ding{55}   & 53.9 & 	47.1 & 	25.4 & 	33.9 	& 66.4 	& 61.4 & 	544.0 & 	76.6 & 	65.1 	& 53.8 \\
    SigLIP2-g~\cite{tschannen2025siglip} & 1.1B    & \ding{55}    &   \ding{55}     & \textbf{57.2} &	48.3 &	28.5 &	37.3 &	78.7 &	75.0 	&671.0 &	\textbf{79.5} &	72.8 &	60.5   \\
    \midrule
    OryxViT~\cite{liu2024oryx} & 0.4B &  \ding{51}      &  \ding{55}      &   56.4    &    48.1   &   26.5    &   39.9    &   78.5   &  79.8  &    660.0   &  78.2     &   72.1      & 60.6  \\
    AIMv2-H~\cite{fini2025multimodal} &    0.7B    & \ding{55}     &   \ding{55}     & 55.2 &	48.2 &	26.8 &	41.8 &	79.1 &	80.1 	&687.0 	&77.8 &	72.5 &	61.1 \\
    InternViT-2.5-6B~\cite{chen2412expanding} &   6.0B     & \ding{55}     &  \ding{55}      & 55.3 &	48.1 &	28.4 &	44.9 &	\textbf{79.9} &	85.7 &	\textbf{757.0} &	78.7 &	\textbf{77.4} 	&63.8 \\
    \midrule
    \rowcolor{red!10} ViQ & 1.3B  &   \ding{51}    &   \ding{51}     & 54.2  & \textbf{49.1}  & \textbf{28.5}  & \textbf{55.3}  & 78.5  & \textbf{88.9}  & 711.0   & 76.7  & 72.8  & \textbf{63.9}  \\
    \bottomrule
    \end{tabular}
  \end{adjustbox}
  \label{tab:multimodal}
\end{table*}

\subsection{Multi-Modal Understanding}
In this subsection, we present experiments on multi-modal understanding to evaluate the effectiveness of our ViQ models in learning superior visual representations. By comparing ViQ with various baseline models, we show that our approach achieves compact visual representations without compromising their quality.

\paragraph{Setups.} We integrate ViQ with large language models of varying scales to evaluate the generalization capability of our vision encoders. To compare with other visual encoders, as ViQ supports native-resolution perception, we adapt the LLaVA-NeXT~\cite{liu2024llavanext}'s ``any resolution" training pipeline for other visual encoders that operate at fixed resolutions to ensure a fair comparison. For evaluation, we adopt representative visual understanding benchmarks, including general multi-modal benchmarks MMStar~\cite{chen2024mmstar}, MMMU~\cite{yue2024mmmu}, world knowledge-relevant benchmarks SimpleVQA~\cite{cheng2025simplevqa}, InfoVQA~\cite{mathew2022infographicvqa}, text and doc recognition benchmarks including TextVQA~\cite{singh2019textvqa}, DocVQA~\cite{mathew2021docvqa}, OCRBench~\cite{liu2023ocrbench}, chart and scientific recognition benchmarks including AI2D~\cite{kembhavi2016ai2d} and ChartQA~\cite{masry2022chartqa}. The comprehensive benchmarks cover a broad range of visual skills, including basic perception, diagram interpretation, OCR, and visual reasoning. Each experiment is conducted on a fixed dataset of 2000K samples drawn from LLaVA-OneVision~\cite{li2024llavaov} to maintain consistency and fairness. We employ LMMs-Eval~\cite{li2024lmms} as our evaluation toolkit to ensure reproducible results.

\paragraph{Visual Encoder Baselines.} To enable a comprehensive comparison with existing visual encoders, we carefully select a diverse set of models across different categories. For general multi-modal visual encoders, we include widely used models such as OpenAI CLIP~\cite{radford2021clip}, SigLIP2~\cite{tschannen2025siglip}, and DINOv2~\cite{oquab2023dinov2}. For visual encoders specialized optimized for mutli-modal data and tasks, we incorporate AIMv2~\cite{fini2025multimodal}, OryxViT~\cite{liu2024oryx}, and InternViT~\cite{zhu2025internvl3} (including 300M and 6B variants). In addition, we evaluate quantized visual encoders, including QLIP~\cite{zhao2025qlip} and UniTok~\cite{ma2025unitok}, to ensure a thorough and balanced comparison.

\paragraph{Results.} As shown in Table~\ref{tab:multimodal}, ViQ achieves competitive performance compared to other visual encoders. Despite being quantized, ViQ matches or surpasses representative general-purpose encoders on the aggregated score, primarily due to its native-resolution perception capability and quantization-aware training strategy, which together preserve strong visual perception throughout quantization. Compared to multimodal encoders that are specifically trained for visual understanding, ViQ remains highly competitive overall and is particularly strong on text- and document-centric tasks. We note, however, that the advantage is not uniform: on certain detail-intensive benchmarks such as OCRBench, ViQ still trails some continuous encoders with fewer parameters, which we attribute to the inherent loss of high-frequency details when the continuous feature space is aggressively compressed into discrete codes. Moreover, ViQ consistently outperforms previous quantized encoders by a large margin, which typically suffer from severe degradation after quantization, highlighting the effectiveness of our proximal representation learning and quantization training strategy in preserving visual quality.

We further observe that ViQ's gains are most pronounced on OCR, document, and infographic understanding tasks. We believe this is because general benchmarks depend more heavily on the backbone LLM's knowledge and reasoning ability, whereas OCR-, document-, and chart-oriented tasks require precise low-level visual details and thus more directly reflect the capability of the visual encoder. From this perspective, the most meaningful contribution of ViQ is that it preserves such fine-grained understanding even after aggressively compressing continuous images into discrete tokens. The residual gap on the most detail-intensive tasks is a systemic property of discrete tokenization rather than a flaw specific to ViQ, and could be further narrowed by orthogonal directions such as multi-scale or residual quantization and incorporating specialized document data. Overall, ViQ serves as an efficient visual encoder that combines high compression with strong perceptual capability, making it well-suited for multimodal understanding tasks.

\subsection{Efficiency}

\begin{figure}[t]
\centering
\includegraphics[width=0.7\linewidth]{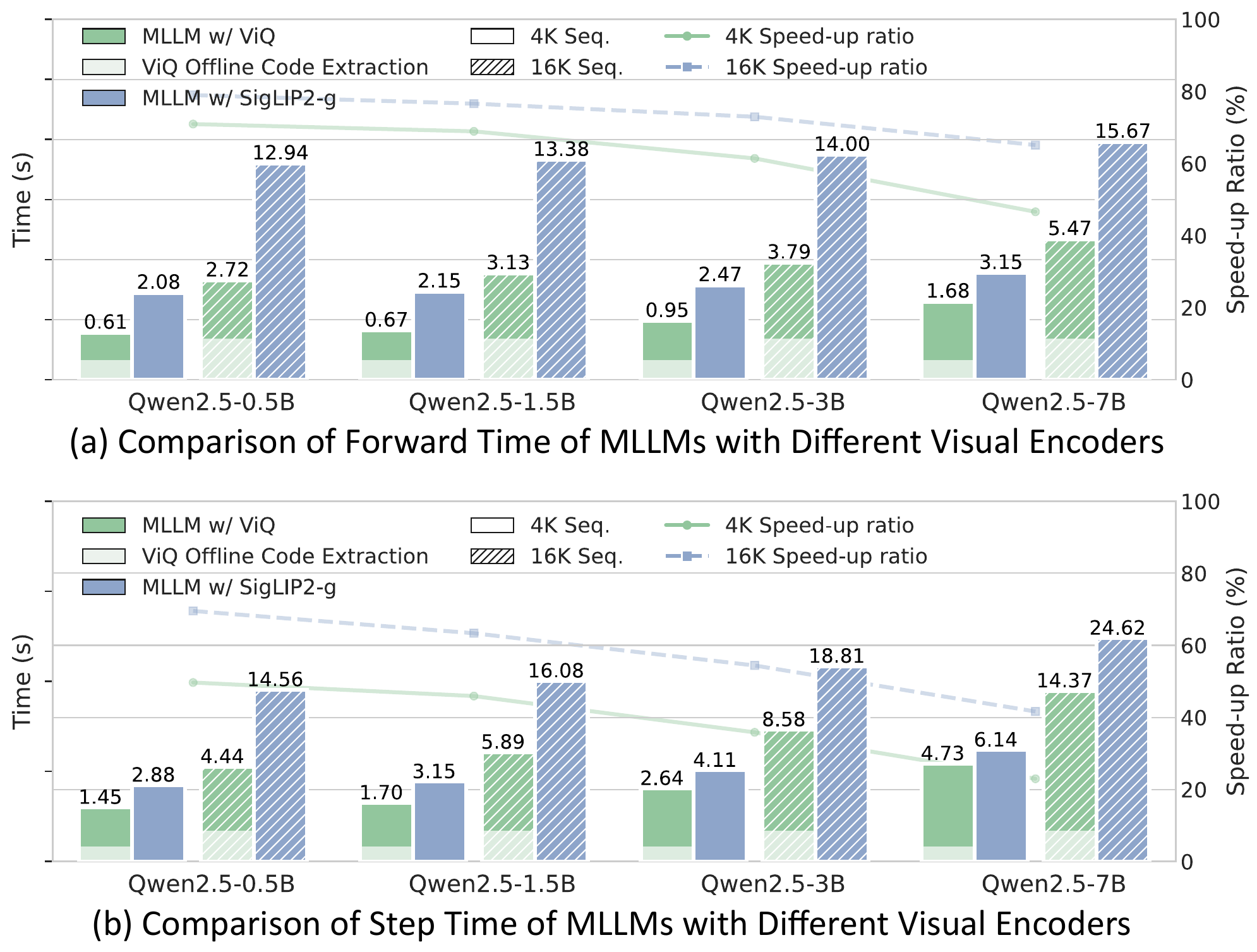}
   \caption{ \small \textbf{Comparisons on Training Efficiency Across Different Visual Encoders. } We conduct the experiments to compare the efficiency of ViQ and SigLIP2-g on the 4k and 16k training.}
   \label{fig:acceleration}
\end{figure}

In this section, we demonstrate ViQ's capabilities beyond multi-modal understanding through some experiments.

\subsubsection{Training Speed-Up for VLMs}
\paragrapha{Setup.} We integrate ViQ and the popular SigLIP2-g encoder with a series of Qwen2.5 models for VLM SFT as conducted in LLaVA~\cite{liu2024llava}. All experiments are conducted on a single node to eliminate noise introduced by inter-node communication. We fix the maximum image area to $768^2$, while preserving the original aspect ratio. We cover the 4k and 16k training cases in experiments, where image-text pair are packed into a fixed sequence length for stable training.

For SigLIP2-g, following common VLM training practice, we extract features from a list of images and feed them into the LLM together with processed text token embeddings for the forward pass. For ViQ, we first extract the discrete codes of each image offline. During the training, instead of loading raw images, we load the precomputed ViQ codes and project them into the LLM latent space. We discard the first 100 warm-up steps and report the average time required for a single forward pass and for a full training iteration.

\paragrapha{Results.} As shown in Figure~\ref{fig:acceleration}, we compare the forward time and step time of SigLIP2-g and ViQ across four different Qwen2.5 model sizes, ranging from 0.5B to 7B. For a fair comparison, we also consider the ViQ offline code extraction time. We can see ViQ provides substantial speed-ups across all model sizes. The gains are especially pronounced for smaller LLMs: for the 0.5B model, ViQ accelerates forward time by 70\% and 78\% under the 4k and 16k training settings, respectively. And for larger model, like Qwen2.5 7B, ViQ can achieve a 46\% and 65\% speed-up in forward time under the 4k and 16k settings. Considering a whole iteration step, ViQ offers consistent improvements, exceeding 20\% and 40\% speed-ups in the 4k and 16k settings.

\subsubsection{Storing Any Image as Discrete Codes} 
\begin{figure*}[t]
\centering
\includegraphics[width=\linewidth]{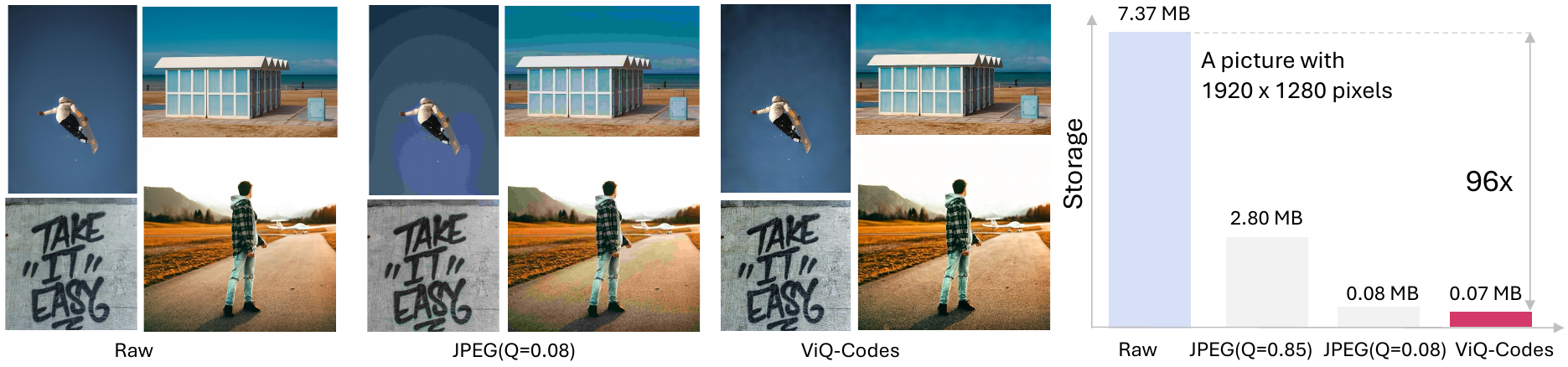}
   \caption{ \small \textbf{Representing images with ViQ.} We show the image compression and visual reconstruction capability of ViQ. ViQ achieves a high-compression-ratio in image storage with high-quality reconstructed images while supporting native-resolution inputs.}
   \label{fig:saving}
\end{figure*}

As ViQ is a quantized visual encoder, it can convert an image at its native resolution into a series of discrete codes, which can then be reconstructed by a decoder. For an image of shape $(H, W)$ with three channels, storing the raw image consumes $\frac{H \times W \times 3 \times 8 \text{bits}}{8} = 3HW$ bytes on disk. In contrast, ViQ can translate an image with $(H, W)$  into $\frac{H\times W}{64}$ codes, each ranging from 0 to 64,000, which can be represented using an unsigned 16-bit integer. As a result, storing the ViQ codes requires $\frac{\frac{H\times W}{64} \times 16 \text{bits}}{8} = \frac{HW}{32}$ bytes, which is only $\frac{1}{96}$ of the raw image size. This corresponds to a very low target bitrate: matching the same compression ratio with JPEG would require an aggressive quality setting (e.g., $Q\approx0.08$) that noticeably degrades image quality, whereas ViQ preserves substantially better reconstruction at an equivalent ratio. We therefore compare ViQ against JPEG at a comparable bitrate, and report the compression ratios together with reconstructed visual samples in Figure~\ref{fig:saving}.

\begin{table}[t]
\centering
\footnotesize
\caption{\small \textbf{Comparison of reconstruction quality on the 256 $\times$ 256 ImageNet-1K validation set.} "Und." indicates whether the tokenizer is optimized for understanding tasks. "*" means reproduced results with the official weights and the same evaluation code as our model.}\vspace{10pt}
\adjustbox{width=\linewidth}{
\label{tab:comparison_vae}
\begin{tabular}{lccccc}
\toprule
\textbf{Method} & ~~~~~~~~\textbf{Und.}~~~~~~~~ & ~~~~~~~~\textbf{\#Token}~~~~~~~~ & ~~~~~~~~\textbf{PSNR$\uparrow$}~~~~~~~~ & ~~~~~~~~\textbf{SSIM$\uparrow$}~~~~~~~~ & ~~~~~~~~\textbf{rFID$\downarrow$}~~~~~~~~ \\
\midrule
\multicolumn{6}{l}{\emph{Continuous Tokenizer}} \\
\midrule
SD-VAE 3 & \ding{55} & 32$\times$32 & 31.29 & 0.87 & 0.20 \\
FLUX-VAE~\cite{laurenccon2024matters} & \ding{55} & 32$\times$32 & 32.74 & 0.92 & 0.18 \\
Qwen-Image~\cite{wu2025qwenimage} & \ding{55} & 32$\times$32 & 32.18 & 0.90 & 1.46 \\
Cosmos-CI~\cite{agarwal2025cosmos} & \ding{55} & 16$\times$16 & 25.07 & 0.70 & 0.96 \\
 Wan2.2~\cite{wan2025wan} & \ding{55} & 16$\times$16 & 31.25 & 0.88 & 0.75 \\
\midrule
\multicolumn{6}{l}{\emph{Discrete Tokenizer}} \\
\midrule
Cosmos-DI~\cite{agarwal2025cosmos} & \ding{55} & 16$\times$16 & 19.98 & 0.54 & 4.40 \\
Show-o~\cite{xie2024show} & \ding{55} & 16$\times$16 & 21.34 & 0.59 & 3.50 \\
LlamaGen~\cite{sun2024autoregressive} & \ding{55} & 16$\times$16 & 20.65 & 0.54 & 2.47 \\
MUSE-VL~\cite{xie2025muse} & \ding{55} & 16$\times$16 & 20.14 & 0.65 & 2.26 \\
Open-MAGVIT2~\cite{luo2024open} & \ding{55} & 16$\times$16 & 22.70 & 0.64 & 1.67 \\
QLIP-B~\cite{zhao2025qlip} & \ding{51} & 16$\times$16 & \underline{23.16} & 0.63 & 3.21 \\
UniTok$*$~\cite{ma2025unitok} & \ding{51} & 16$\times$16 & \textbf{25.32} & \textbf{0.77} & \textbf{0.37} \\
\rowcolor{red!10} ViQ & \ding{51} & 16$\times$16 & 22.73 & \underline{0.66} & \underline{0.62} \\
\bottomrule
\end{tabular}}
\end{table}

\subsection{Image Reconstruction Experiments}
\paragraph{Setups.} For image reconstruction tasks, we train the visual decoder with the fixed pre-trained ViQ model. The overall regularization loss for VAE is defined as:
\begin{equation}
    \mathcal{L}_{\text{VAE}} = \mathcal{L}_{\text{KL}} + \mathcal{L}_{\text{MSE}} + \mathcal{L}_{\text{LPIPS}} + \mathcal{L}_{\text{GAN}} + \lambda_{\text{REPA}} \cdot \mathcal{L}_{\text{REPA}}
\end{equation}
where $\mathcal{L}_{\text{KL}}$ denotes the Kullback-Leibler divergence, $\mathcal{L}_{\text{MSE}}$ represents the Mean Squared Error, $\mathcal{L}_{\text{LPIPS}}$ corresponds to the Learned Perceptual Image Patch Similarity for pixel reconstruction, and $\mathcal{L}_{\text{GAN}}$ is the adversarial loss from the Generative Adversarial Network~\cite{goodfellow2020generative}. To enhance semantic alignment, we leverage DINOv2~\cite{oquab2023dinov2} as an external feature extractor and apply an alignment loss $\mathcal{L}_{\text{REPA}}$ to the ViQ preprocess layer preceding the decoder head within our ViQ architecture. The alignment loss coefficient is empirically set to $\lambda_{\text{REPA}} = 1.5$ for the DINOv2 head.
For model optimization, we employ the AdamW optimizer~\cite{loshchilov2017decoupled, kingma2017adammethodstochasticoptimization} with a constant learning rate of $6\times10^{-4}$ and a global batch size of 4096. Training stability is ensured through gradient clipping and the application of an Exponential Moving Average (EMA) to the generative model parameters. All models are trained for 50,000 steps with a linear learning rate warmup over the initial 5,000 steps. We utilize mixed-precision training with the BF16 format and unfreeze the VAE preprocess layer during training.

\paragraph{Results.}
We comprehensively evaluate the reconstruction quality of various tokenizers on the 256$\times$256 ImageNet-1K validation set, as summarized in Table~\ref{tab:comparison_vae}.
Among discrete tokenizers, ViQ delivers highly competitive reconstruction quality, attaining an SSIM of 0.66 and an rFID of 0.62 (second only to UniTok), while remaining comparable to QLIP-B in PSNR (22.73 vs. 23.16). We note that UniTok reports stronger raw reconstruction metrics (e.g., PSNR 25.32), which we attribute to its direct reconstruction objective combined with contrastive supervision. However, as shown in Table~\ref{tab:multimodal}, this type of supervision substantially degrades vision-language alignment and thus leads to sub-optimal multi-modal understanding. By contrast, our core objective with ViQ is to deliver a more balanced tokenizer that maintains highly competitive low-level reconstruction while excelling in high-level semantic understanding.
This balance indicates that a well-optimized pixel-based decoder remains a powerful and efficient choice for building a discrete visual tokenizer that serves both generative and discriminative purposes.
We also acknowledge that, as a discrete tokenizer, ViQ inevitably compromises some fine-grained reconstruction detail relative to continuous tokenizers, as the continuous visual feature space is drastically compressed; nevertheless, it considerably narrows this gap and offers a favorable trade-off between reconstruction fidelity and semantic understanding.

\begin{table*}[t]
\vspace{-.2em}
\centering
\renewcommand{\arraystretch}{1.05}
\setlength{\tabcolsep}{4pt}

\subfloat[
\textbf{Proximal Representations}. Gradually regularizing the latent space from continuous to quantized helps.
\label{tab:abla-1}
]{
\begin{minipage}{0.31\linewidth}
\centering
\adjustbox{width=\linewidth}{
\begin{tabular}{cc}
Process & avg.(2-2) \\
\hline
Continuous $\to$ SimVQ & 60.9 \\
Continuous $\to$ BN + $l_{2}$ $\to$ SimVQ & 66.6 \\
Continuous $\to$ BN + $l_{2}$ $\to$ FSQ & 67.9 \\
\rowcolor{red!10} Continuous $\to$ BN + $l_{\infty}$ $\to$ FSQ & 68.7 \\
\end{tabular}}
\end{minipage}
}
\hfill
\subfloat[
\textbf{Bottleneck size}. The bottleneck width can be made significantly narrower than the 1536-dimensional one in SigLIP2-g.
\label{tab:abla-2}
]{
\begin{minipage}{0.31\linewidth}
\centering
\adjustbox{width=\linewidth}{
\begin{tabular}{cc}
~~~~~~~~~Width~~~~~~~~~ & ~~~~~~~~~avg.(2-1)~~~~~~~~~ \\
\hline
32  & 68.4 \\
\rowcolor{red!10} 128  & 69.1 \\
512 & 68.8 \\
1536 & 69.3 \\
\end{tabular}}
\end{minipage}
}
\hfill
\subfloat[
\textbf{VQ and Codebook size}.
The VQ codebook saturates at a size of $2^{17}$.
\label{tab:abla-3}
]{
\begin{minipage}{0.31\linewidth}
\centering
\adjustbox{width=\linewidth}{
\begin{tabular}{cc}
~~~~~~~~~Size~~~~~~~~~ & ~~~~~~~~~avg.(2-2)~~~~~~~~~ \\
\hline
\rowcolor{red!10} FSQ + 64,000 & 68.7\\
FSQ + 128,000 & 68.3\\
SimVQ + $2^{15}$ & 66.5\\
SimVQ + $2^{17}$ & 65.6\\
\end{tabular}}
\end{minipage}
}

\vspace{1em}

\subfloat[
\textbf{Position encoding for Quantization}. position information is important.
\label{tab:abla-4}
]{
\begin{minipage}{0.31\linewidth}
\centering
\adjustbox{width=\linewidth}{
\begin{tabular}{cc}
Design & avg.(2-2) \\
\hline
no Position information injected & 65.3\\
\rowcolor{red!10} RoPE with Attention & 68.7 \\
Learnable Pos Emb & 65.7 \\
\end{tabular}}
\end{minipage}
}
\hfill
\subfloat[
\textbf{Loss Combination}. three type of loss.
\label{tab:abla-5}
]{
\begin{minipage}{0.31\linewidth}
\centering
\adjustbox{width=\linewidth}{
\begin{tabular}{ccccc}
~Case~ & Self-Distill & Text & Recon & avg.(2-2)\\
\hline
A & \ding{55} & \ding{51} & \ding{55} & 61.3 \\
B & \ding{51} & \ding{51} & \ding{55} & 66.8 \\
\rowcolor{red!10} C & \ding{51} & \ding{51} & \ding{51} & 68.7\\
\end{tabular}}
\end{minipage}
}
\hfill
\subfloat[
\textbf{Reconstruction Loss}. Vae Latent loss is efficient.
\label{tab:abla-6}
]{
\begin{minipage}{0.32\linewidth}
\centering
\adjustbox{width=\linewidth}{
\begin{tabular}{ccc}
~~~~Loss Type~~~~ & ~Time Cost~ & ~~avg.(2-2)~~ \\
\hline
none & 1x & 66.8 \\
MSE + LPIPS & 2.3x & 67.0 \\
DiT (unfrozen) & 4x & 65.8 \\
DiT (frozen) & 4x & 67.6 \\
\rowcolor{red!10} Vae Latent Loss& 1.3x & 68.7 \\
\end{tabular}}
\end{minipage}
}

\vspace{10pt}
\caption{\textbf{ViQ ablation experiments}. We conduct a thorough ablation study of ViQ, examining proximal representations, architecture design, and loss combinations. We report the average metrics on MMStar, MMMU, OCRBench, among a total of eight benchmarks, using a fixed SFT training setup on Qwen2.5-3B. (2-1) and (2-2) refer to the training stage of the test model. Default settings are marked in \colorbox{red!10}{red}.}
\label{tab:ablations}
\end{table*}

\subsection{Ablation Studies}

\paragrapha{Proximal Representations}
In Tables~\ref{tab:abla-1} and \ref{tab:abla-2}, we show that initializing the model with intermediate proximal representations can substantially improve the final performance.
In Table~\ref{tab:abla-1}, we can see that optimizing a continuous feature space into a discrete one leads to a severe performance drop. We further investigate the effect of introducing a bottleneck, a bottleneck with L2 normalization, and a bottleneck with L$_\infty$ normalization. The results highlight two key observations: first, a gradually regularized latent space is more suitable for quantization; second, a more appropriate regularization method for quantization (i.e., L$_\infty$) provides additional benefits. Furthermore, Table~\ref{tab:abla-2} shows that, when optimized properly, the bottleneck does not necessarily degrade the final performance.

\paragrapha{Quantization Designs}
We compared the performance of commonly used methods such as SimVQ~\cite{zhu2025addressing} and FSQ~\cite{mentzerfinite} under different codebook sizes, as shown in Table~\ref{tab:abla-3}. We conduct experiments on different position encoding methods for ViQ, summarized in Table~\ref{tab:abla-4}. We observe that FSQ~\cite{mentzerfinite}, a non-optimized vector quantization method, outperforms SimVQ~\cite{zhu2025addressing}, which requires learning a codebook. We also experimented with LFQ~\cite{yulanguage}, vanilla VQ~\cite{gersho2012vector}, and IBQ~\cite{shi2025scalable}, and found consistent conclusions: in our setting, VQ methods without a learnable vocabulary tend to perform better. In addition, we find that a vocabulary size around 60,000 is good. Increasing the codebook size will reduce the utilization rate of learnable codebooks, which in turn degrades performance, whereas this issue is much less pronounced for non-learnable VQ methods. Table~\ref{tab:abla-4} further shows that injecting positional encoding significantly enhances the representational ability of VQ. We conduct experiments on both learnable positional embeddings and RoPE-2D, and observe that RoPE leads to a substantial improvement, whereas learnable positional embeddings offer limited gains. This is because learnable positional embeddings increase the optimization difficulty of the VQ module.

\paragrapha{Loss Combination. }
We further study the loss composition used in our training pipeline. As shown in Table~\ref{tab:abla-5}. When keep only the text loss, it leads to a substantial performance drop (Case A). In Case B, add self distillation loss with significant enhance the performace of model. While we can see some unsatisfactory performance on tasks such as OCR and Chart. By introducing a reconstruction loss (Case C), the performance notably improves, with most gains coming from detail-intensive tasks such as OCR and Chart, while the gains for captioning and VQA are relatively limited. Table~\ref{tab:abla-6} studies different formulations of the reconstruction loss. We adopt a more efficient and simple approach: a VAE latent reconstruction loss, where the model predicts the latent representation of a pretrained VAE network. This achieves effective optimization while maintaining training efficiency. We further analyze why the alternative formulations are less effective. For the DiT-based objective, jointly training an \emph{unfrozen} DiT module lets the generative branch absorb part of the representational capacity of the encoder, which degrades the discriminative features and even drops below the no-reconstruction baseline (65.8); freezing the DiT alleviates this and restores the score to 67.6, yet it still trails the VAE latent loss due to the lack of fine-tuning flexibility. For pixel-level objectives, MSE and LPIPS mainly enforce rigid pixel-wise fitting, which tends to compete with and erode the high-level semantic features that are crucial for multi-modal understanding. By contrast, supervising in the latent space of a pretrained VAE strikes a better balance: it guides low-level detail reconstruction while preserving semantic integrity, and does so at a much lower computational cost ($1.3\times$).

\subsection{Limitations}
Although ViQ demonstrates highly competitive performance in unifying visual and language representations into a cohesive discrete framework, there are a few general limitations to consider. While the study thoroughly validates ViQ's efficiency and effectiveness across large language models ranging from 0.5B to 7B parameters, its integration and synergistic effects with massively larger foundation models (e.g., 70B parameters and beyond) remain an area for future empirical exploration. Moreover, like most advanced multimodal models, the robustness of the learned proximal representations relies on the quality and diversity of the large-scale language-image pre-training data, meaning that inherent data biases could marginally affect the model's zero-shot generalization in highly specialized or extreme out-of-domain scenarios.

\section{Conclusion}
\label{sec:conclusion}

In this work, we introduced ViQ, a quantized multimodal encoder that unifies visual and language representations at native resolution. We designed a two-stage training pipeline to reduce the information losses of learning discrete visual representations with text-aligned pretraining and feature discretization. Through carefully designed architectures and progressive training techniques that minimize information loss, ViQ achieves competitive performance in multimodal tasks compared to both existing continuous and discrete visual encoders. We believe our work not only enhances the quality and efficiency of visual representations in multimodal learning, but also offers a viable pathway toward unifying vision and language within a shared discrete framework.

\newpage

\renewcommand{\refname}{References}
\renewcommand{\bibname}{References}
\renewcommand{\bibsection}{\section*{\raggedright \Large References}}
\bibliographystyle{plainnat}
\bibliography{references}

@String(CVPR= {IEEE Conf. Comput. Vis. Pattern Recog.})

@String(ECCV= {Eur. Conf. Comput. Vis.})

@String(CVPR  = {CVPR})

@String(ECCV  = {ECCV})

@article{qwen2vl,
  title={Qwen2-VL: To See the World More Clearly},
  author={QwenTeam},
  journal = {Wwen Blog},
  url = {https://qwenlm.github.io/blog/qwen2-vl/},
  year={2024}
}

@article{li2024llavaov,
  title={LLaVA-OneVision: Easy Visual Task Transfer},
  author={Li, Bo and Zhang, Yuanhan and Guo, Dong and Zhang, Renrui and Li, Feng and Zhang, Hao and Zhang, Kaichen and Li, Yanwei and Liu, Ziwei and Li, Chunyuan},
  journal={arXiv preprint arXiv:2408.03326},
  year={2024}
}

@inproceedings{chen2024internvl,
  title={Internvl: Scaling up vision foundation models and aligning for generic visual-linguistic tasks},
  author={Chen, Zhe and Wu, Jiannan and Wang, Wenhai and Su, Weijie and Chen, Guo and Xing, Sen and Zhong, Muyan and Zhang, Qinglong and Zhu, Xizhou and Lu, Lewei and others},
  booktitle={CVPR},
  pages={24185--24198},
  year={2024}
}

@article{xie2024show,
  title={Show-o: One Single Transformer to Unify Multimodal Understanding and Generation},
  author={Xie, Jinheng and Mao, Weijia and Bai, Zechen and Zhang, David Junhao and Wang, Weihao and Lin, Kevin Qinghong and Gu, Yuchao and Chen, Zhijie and Yang, Zhenheng and Shou, Mike Zheng},
  journal={arXiv preprint arXiv:2408.12528},
  year={2024}
}

@article{liu2024llava,
  title={Visual instruction tuning},
  author={Liu, Haotian and Li, Chunyuan and Wu, Qingyang and Lee, Yong Jae},
  journal={NeurIPS},
  volume={36},
  year={2024}
}

@misc{liu2024llavanext,
  title={Llava-next: Improved reasoning, ocr, and world knowledge},
  author={Liu, Haotian and Li, Chunyuan and Li, Yuheng and Li, Bo and Zhang, Yuanhan and Shen, Sheng and Lee, Yong Jae},
  year={2024}
}

@inproceedings{zhai2023siglip,
  title={Sigmoid loss for language image pre-training},
  author={Zhai, Xiaohua and Mustafa, Basil and Kolesnikov, Alexander and Beyer, Lucas},
  booktitle={Proceedings of the IEEE/CVF International Conference on Computer Vision},
  pages={11975--11986},
  year={2023}
}

@inproceedings{radford2021clip,
  title={Learning transferable visual models from natural language supervision},
  author={Radford, Alec and Kim, Jong Wook and Hallacy, Chris and Ramesh, Aditya and Goh, Gabriel and Agarwal, Sandhini and Sastry, Girish and Askell, Amanda and Mishkin, Pamela and Clark, Jack and others},
  booktitle={ICML},
  pages={8748--8763},
  year={2021},
  organization={PMLR}
}

@article{dehghani2024navit,
  title={Patch n’pack: Navit, a vision transformer for any aspect ratio and resolution},
  author={Dehghani, Mostafa and Mustafa, Basil and Djolonga, Josip and Heek, Jonathan and Minderer, Matthias and Caron, Mathilde and Steiner, Andreas and Puigcerver, Joan and Geirhos, Robert and Alabdulmohsin, Ibrahim M and others},
  journal={NeurIPS},
  volume={36},
  year={2024}
}

@article{laurenccon2024matters,
  title={What matters when building vision-language models?},
  author={Lauren{\c{c}}on, Hugo and Tronchon, L{\'e}o and Cord, Matthieu and Sanh, Victor},
  journal={arXiv preprint arXiv:2405.02246},
  year={2024}
}

@inproceedings{yue2024mmmu,
  title={Mmmu: A massive multi-discipline multimodal understanding and reasoning benchmark for expert agi},
  author={Yue, Xiang and Ni, Yuansheng and Zhang, Kai and Zheng, Tianyu and Liu, Ruoqi and Zhang, Ge and Stevens, Samuel and Jiang, Dongfu and Ren, Weiming and Sun, Yuxuan and others},
  booktitle={Proceedings of the IEEE/CVF Conference on Computer Vision and Pattern Recognition},
  pages={9556--9567},
  year={2024}
}

@inproceedings{mathew2021docvqa,
  title={Docvqa: A dataset for vqa on document images},
  author={Mathew, Minesh and Karatzas, Dimosthenis and Jawahar, CV},
  booktitle={Proceedings of the IEEE/CVF winter conference on applications of computer vision},
  pages={2200--2209},
  year={2021}
}

@article{liu2023ocrbench,
  title={On the hidden mystery of ocr in large multimodal models},
  author={Liu, Yuliang and Li, Zhang and Yang, Biao and Li, Chunyuan and Yin, Xucheng and Liu, Cheng-lin and Jin, Lianwen and Bai, Xiang},
  journal={arXiv preprint arXiv:2305.07895},
  year={2023}
}

@inproceedings{kembhavi2016ai2d,
  title={A diagram is worth a dozen images},
  author={Kembhavi, Aniruddha and Salvato, Mike and Kolve, Eric and Seo, Minjoon and Hajishirzi, Hannaneh and Farhadi, Ali},
  booktitle={ECCV},
  pages={235--251},
  year={2016},
  organization={Springer}
}

@inproceedings{singh2019textvqa,
  title={Towards vqa models that can read},
  author={Singh, Amanpreet and Natarajan, Vivek and Shah, Meet and Jiang, Yu and Chen, Xinlei and Batra, Dhruv and Parikh, Devi and Rohrbach, Marcus},
  booktitle={Proceedings of the IEEE/CVF conference on computer vision and pattern recognition},
  pages={8317--8326},
  year={2019}
}

@article{liu2024oryx,
  title={Oryx MLLM: On-Demand Spatial-Temporal Understanding at Arbitrary Resolution},
  author={Liu, Zuyan and Dong, Yuhao and Liu, Ziwei and Hu, Winston and Lu, Jiwen and Rao, Yongming},
  journal={arXiv preprint arXiv:2409.12961},
  year={2024}
}

@misc{Qwen2.5-VL,
    title = {Qwen2.5-VL},
    url = {https://qwenlm.github.io/blog/qwen2.5-vl/},
    author = {Qwen Team},
    month = {January},
    year = {2025}
}

@article{chen2024mmstar,
  title={Are We on the Right Way for Evaluating Large Vision-Language Models?},
  author={Chen, Lin and Li, Jinsong and Dong, Xiaoyi and Zhang, Pan and Zang, Yuhang and Chen, Zehui and Duan, Haodong and Wang, Jiaqi and Qiao, Yu and Lin, Dahua and others},
  journal={arXiv preprint arXiv:2403.20330},
  year={2024}
}

@article{masry2022chartqa,
  title={Chartqa: A benchmark for question answering about charts with visual and logical reasoning},
  author={Masry, Ahmed and Long, Do Xuan and Tan, Jia Qing and Joty, Shafiq and Hoque, Enamul},
  journal={arXiv preprint arXiv:2203.10244},
  year={2022}
}

@article{zhu2025internvl3,
  title={Internvl3: Exploring advanced training and test-time recipes for open-source multimodal models},
  author={Zhu, Jinguo and Wang, Weiyun and Chen, Zhe and Liu, Zhaoyang and Ye, Shenglong and Gu, Lixin and Tian, Hao and Duan, Yuchen and Su, Weijie and Shao, Jie and others},
  journal={arXiv preprint arXiv:2504.10479},
  year={2025}
}

@article{chen2412expanding,
  title={Expanding performance boundaries of open-source multimodal models with model, data, and test-time scaling, 2025},
  author={Chen, Zhe and Wang, Weiyun and Cao, Yue and Liu, Yangzhou and Gao, Zhangwei and Cui, Erfei and Zhu, Jinguo and Ye, Shenglong and Tian, Hao and Liu, Zhaoyang and others},
  journal={URL https://arxiv. org/abs/2412.05271}
}

@article{oquab2023dinov2,
  title={Dinov2: Learning robust visual features without supervision},
  author={Oquab, Maxime and Darcet, Timoth{\'e}e and Moutakanni, Th{\'e}o and Vo, Huy and Szafraniec, Marc and Khalidov, Vasil and Fernandez, Pierre and Haziza, Daniel and Massa, Francisco and El-Nouby, Alaaeldin and others},
  journal={arXiv preprint arXiv:2304.07193},
  year={2023}
}

@inproceedings{fini2025multimodal,
  title={Multimodal autoregressive pre-training of large vision encoders},
  author={Fini, Enrico and Shukor, Mustafa and Li, Xiujun and Dufter, Philipp and Klein, Michal and Haldimann, David and Aitharaju, Sai and da Costa, Victor G Turrisi and B{\'e}thune, Louis and Gan, Zhe and others},
  booktitle={Proceedings of the Computer Vision and Pattern Recognition Conference},
  pages={9641--9654},
  year={2025}
}

@article{tschannen2025siglip,
  title={Siglip 2: Multilingual vision-language encoders with improved semantic understanding, localization, and dense features},
  author={Tschannen, Michael and Gritsenko, Alexey and Wang, Xiao and Naeem, Muhammad Ferjad and Alabdulmohsin, Ibrahim and Parthasarathy, Nikhil and Evans, Talfan and Beyer, Lucas and Xia, Ye and Mustafa, Basil and others},
  journal={arXiv preprint arXiv:2502.14786},
  year={2025}
}

@article{zhao2025qlip,
  title={QLIP: Text-Aligned Visual Tokenization Unifies Auto-Regressive Multimodal Understanding and Generation},
  author={Zhao, Yue and Xue, Fuzhao and Reed, Scott and Fan, Linxi and Zhu, Yuke and Kautz, Jan and Yu, Zhiding and Kr{\"a}henb{\"u}hl, Philipp and Huang, De-An},
  journal={CoRR},
  year={2025}
}

@article{ma2025unitok,
  title={Unitok: A unified tokenizer for visual generation and understanding},
  author={Ma, Chuofan and Jiang, Yi and Wu, Junfeng and Yang, Jihan and Yu, Xin and Yuan, Zehuan and Peng, Bingyue and Qi, Xiaojuan},
  journal={arXiv preprint arXiv:2502.20321},
  year={2025}
}

@inproceedings{mentzerfinite,
  title={Finite Scalar Quantization: VQ-VAE Made Simple},
  author={Mentzer, Fabian and Minnen, David and Agustsson, Eirikur and Tschannen, Michael},
  booktitle={The Twelfth International Conference on Learning Representations}
}

@book{gersho2012vector,
  title={Vector quantization and signal compression},
  author={Gersho, Allen and Gray, Robert M},
  volume={159},
  year={2012},
  publisher={Springer Science \& Business Media}
}

@inproceedings{chiu2022self,
  title={Self-supervised learning with random-projection quantizer for speech recognition},
  author={Chiu, Chung-Cheng and Qin, James and Zhang, Yu and Yu, Jiahui and Wu, Yonghui},
  booktitle={International Conference on Machine Learning},
  pages={3915--3924},
  year={2022},
  organization={PMLR}
}

@inproceedings{yulanguage,
  title={Language Model Beats Diffusion-Tokenizer is key to visual generation},
  author={Yu, Lijun and Lezama, Jose and Gundavarapu, Nitesh Bharadwaj and Versari, Luca and Sohn, Kihyuk and Minnen, David and Cheng, Yong and Gupta, Agrim and Gu, Xiuye and Hauptmann, Alexander G and others},
  booktitle={The Twelfth International Conference on Learning Representations}
}

@inproceedings{zhaoimage,
  title={Image and Video Tokenization with Binary Spherical Quantization},
  author={Zhao, Yue and Xiong, Yuanjun and Kraehenbuehl, Philipp},
  booktitle={The Thirteenth International Conference on Learning Representations}
}

@inproceedings{esser2021taming,
  title={Taming transformers for high-resolution image synthesis},
  author={Esser, Patrick and Rombach, Robin and Ommer, Bjorn},
  booktitle={Proceedings of the IEEE/CVF conference on computer vision and pattern recognition},
  pages={12873--12883},
  year={2021}
}

@article{van2017neural,
  title={Neural discrete representation learning},
  author={Van Den Oord, Aaron and Vinyals, Oriol and others},
  journal={Advances in neural information processing systems},
  volume={30},
  year={2017}
}

@article{loshchilov2017decoupled,
  title={Decoupled weight decay regularization},
  author={Loshchilov, Ilya and Hutter, Frank},
  journal={arXiv preprint arXiv:1711.05101},
  year={2017}
}

@misc{kingma2017adammethodstochasticoptimization,
      title={Adam: A Method for Stochastic Optimization}, 
      author={Diederik P. Kingma and Jimmy Ba},
      year={2017},
      eprint={1412.6980},
      archivePrefix={arXiv},
      primaryClass={cs.LG},
      url={https://arxiv.org/abs/1412.6980}, 
}

@article{goodfellow2020generative,
  title={Generative adversarial networks},
  author={Goodfellow, Ian and Pouget-Abadie, Jean and Mirza, Mehdi and Xu, Bing and Warde-Farley, David and Ozair, Sherjil and Courville, Aaron and Bengio, Yoshua},
  journal={Communications of the ACM},
  volume={63},
  number={11},
  pages={139--144},
  year={2020},
  publisher={ACM New York, NY, USA}
}

@article{luo2024open,
  title={Open-magvit2: An open-source project toward democratizing auto-regressive visual generation},
  author={Luo, Zhuoyan and Shi, Fengyuan and Ge, Yixiao and Yang, Yujiu and Wang, Limin and Shan, Ying},
  journal={arXiv preprint arXiv:2409.04410},
  year={2024}
}

@inproceedings{xie2025muse,
  title={Muse-vl: Modeling unified vlm through semantic discrete encoding},
  author={Xie, Rongchang and Du, Chen and Song, Ping and Liu, Chang},
  booktitle={Proceedings of the IEEE/CVF International Conference on Computer Vision},
  pages={24135--24146},
  year={2025}
}

@article{sun2024autoregressive,
  title={Autoregressive model beats diffusion: Llama for scalable image generation},
  author={Sun, Peize and Jiang, Yi and Chen, Shoufa and Zhang, Shilong and Peng, Bingyue and Luo, Ping and Yuan, Zehuan},
  journal={arXiv preprint arXiv:2406.06525},
  year={2024}
}

@article{agarwal2025cosmos,
  title={Cosmos world foundation model platform for physical ai},
  author={Agarwal, Niket and Ali, Arslan and Bala, Maciej and Balaji, Yogesh and Barker, Erik and Cai, Tiffany and Chattopadhyay, Prithvijit and Chen, Yongxin and Cui, Yin and Ding, Yifan and others},
  journal={arXiv preprint arXiv:2501.03575},
  year={2025}
}

@article{wan2025wan,
  title={Wan: Open and advanced large-scale video generative models},
  author={Wan, Team and Wang, Ang and Ai, Baole and Wen, Bin and Mao, Chaojie and Xie, Chen-Wei and Chen, Di and Yu, Feiwu and Zhao, Haiming and Yang, Jianxiao and others},
  journal={arXiv preprint arXiv:2503.20314},
  year={2025}
}

@article{wu2025qwenimage,
  title={Qwen-image technical report},
  author={Wu, Chenfei and Li, Jiahao and Zhou, Jingren and Lin, Junyang and Gao, Kaiyuan and Yan, Kun and Yin, Sheng-ming and Bai, Shuai and Xu, Xiao and Chen, Yilei and others},
  journal={arXiv preprint arXiv:2508.02324},
  year={2025}
}

@article{fang2023dfn,
  title={Data filtering networks},
  author={Fang, Alex and Jose, Albin Madappally and Jain, Amit and Schmidt, Ludwig and Toshev, Alexander and Shankar, Vaishaal},
  journal={arXiv preprint arXiv:2309.17425},
  year={2023}
}

@inproceedings{cheng2025simplevqa,
  title={Simplevqa: Multimodal factuality evaluation for multimodal large language models},
  author={Cheng, Xianfu and Zhang, Wei and Zhang, Shiwei and Yang, Jian and Guan, Xiangyuan and Wu, Xianjie and Li, Xiang and Zhang, Ge and Liu, Jiaheng and Mai, Yuying and others},
  booktitle={Proceedings of the IEEE/CVF International Conference on Computer Vision},
  pages={4637--4646},
  year={2025}
}

@inproceedings{mathew2022infographicvqa,
  title={Infographicvqa},
  author={Mathew, Minesh and Bagal, Viraj and Tito, Rub{\`e}n and Karatzas, Dimosthenis and Valveny, Ernest and Jawahar, CV},
  booktitle={Proceedings of the IEEE/CVF Winter Conference on Applications of Computer Vision},
  pages={1697--1706},
  year={2022}
}

@misc{li2024lmms,
  title={Lmms-eval: Accelerating the development of large multimoal models},
  author={Li, Bo and Zhang, Peiyuan and Zhang, Kaichen and Pu, Fanyi and Du, Xinrun and Dong, Yuhao and Liu, Haotian and Zhang, Yuanhan and Zhang, Ge and Li, Chunyuan and others},
  year={2024},
  publisher={March}
}

@article{yin2025sail,
  title={Sail-vl2 technical report},
  author={Yin, Weijie and Ye, Yongjie and Shu, Fangxun and Liao, Yue and Kang, Zijian and Dong, Hongyuan and Yu, Haiyang and Yang, Dingkang and Wang, Jiacong and Wang, Han and others},
  journal={arXiv preprint arXiv:2509.14033},
  year={2025}
}

@article{guo2025seed1,
  title={Seed1. 5-vl technical report},
  author={Guo, Dong and Wu, Faming and Zhu, Feida and Leng, Fuxing and Shi, Guang and Chen, Haobin and Fan, Haoqi and Wang, Jian and Jiang, Jianyu and Wang, Jiawei and others},
  journal={arXiv preprint arXiv:2505.07062},
  year={2025}
}

@article{team2025kimi,
  title={Kimi-vl technical report},
  author={Team, Kimi and Du, Angang and Yin, Bohong and Xing, Bowei and Qu, Bowen and Wang, Bowen and Chen, Cheng and Zhang, Chenlin and Du, Chenzhuang and Wei, Chu and others},
  journal={arXiv preprint arXiv:2504.07491},
  year={2025}
}

@inproceedings{zhu2025addressing,
  title={Addressing representation collapse in vector quantized models with one linear layer},
  author={Zhu, Yongxin and Li, Bocheng and Xin, Yifei and Xia, Zhihua and Xu, Linli},
  booktitle={Proceedings of the IEEE/CVF International Conference on Computer Vision},
  pages={22968--22977},
  year={2025}
}

@inproceedings{shi2025scalable,
  title={Scalable image tokenization with index backpropagation quantization},
  author={Shi, Fengyuan and Luo, Zhuoyan and Ge, Yixiao and Yang, Yujiu and Shan, Ying and Wang, Limin},
  booktitle={Proceedings of the IEEE/CVF International Conference on Computer Vision},
  pages={16037--16046},
  year={2025}
}

@article{su2024roformer,
  title={Roformer: Enhanced transformer with rotary position embedding},
  author={Su, Jianlin and Ahmed, Murtadha and Lu, Yu and Pan, Shengfeng and Bo, Wen and Liu, Yunfeng},
  journal={Neurocomputing},
  volume={568},
  pages={127063},
  year={2024},
  publisher={Elsevier}
}

@misc{meituanlongcatteam2026longcatnextlexicalizingmodalitiesdiscrete,
      title={LongCat-Next: Lexicalizing Modalities as Discrete Tokens}, 
      author={Meituan LongCat Team},
      year={2026},
      eprint={2603.27538},
      archivePrefix={arXiv},
      primaryClass={cs.CV},
      url={https://arxiv.org/abs/2603.27538}, 
}

\newpage
\appendix
\setcounter{section}{0}
\renewcommand{\thesection}{\Alph{section}}
{\large\bfseries Appendix\par}
\addcontentsline{toc}{section}{Appendix}
\vspace{1em}
\section{More Details}
\label{sec:training}
\subsection{Training Details}
\paragrapha{From Fix Resolution to Native Resolution. }
We first train our ViT with a light-weight LLM on approximately 3B vision–language tokens, covering tasks such as captioning, chart understanding, OCR, and general diagram comprehension. In this stage, we constrained the image size to within $384^2$ while preserving each image's native aspect ratio. For training efficiency, we downsampled the vision tokens by a factor of 16 before feeding them into the LLM. After this stage, we increased the image size to 768$^2$ and introduced another $\sim$3B training tokens. During this phase, the vision tokens were downsampled by a factor of 4 before being passed to the LLM for text loss supervision. The learning rate was gradually decayed from 2e-5 to 5e-5 for above two stage, respectively. We used Qwen2.5-0.5B as the LLM backbone and optimized only the LoRA parameters. For the self distillation loss, we use the cosine similarity loss for the global features between original SigLIP2-g and trained ViT wit a Multi-head Attention Pooling Layer. The images for SigLIP2-g are resized to 384$\times$384.

\paragrapha{From Continuous to Quantized. }
In Stage 2-1, we optimized a bottleneck with a dimension of 128, where newly added parameters were initialized with a learning rate of $1\times 10^{-4}$, while all other parameters used a learning rate of $5\times 10^{-5}$, half of the initial value. The learning rate was gradually decayed using a cosine scheduler, ultimately reaching $1\times 10^{-5}$ and $5\times 10^{-6}$ for different parameter groups. The data source used in Stage 1 was identical to the one mentioned earlier, and we trained on 1B vision-language tokens with a resolution of 768 pixels. The bottleneck was optimized during this phase. For the reconstruction branch, we passed the final vision feature through a head designed for VAE latent feature prediction. This head was structured as a 3-layer MHSA and culminated in a convolutional layer for upsampling. Additionally, vision features underwent a 2x2 pooling operation before being fed into the Large Language Model. During this process, we introduced constraints on the bottleneck's 128-dimensional features, such as an $L_{\infty}$ norm regularization. In Stage 2-2, we further refined the bottleneck from Stage 1. The previously applied non-parameteric constraints like the $L_{\infty}$ regularization were replaced with the FSQ module. This module incorporates a quantization mechanism along the 6-dimensional feature space, followed by fully connected layers before and after quantization, along with an attention layer that adds Rotary Positional Embedding~\cite{su2024roformer} information. In this stage, we utilized learning rate of $5\times 10^{-5}$ across all components and trained on approximately 30B vision-language tokens.

\section{Size-Matched Comparison}
To isolate the benefit of our quantization approach from the effect of model scale, we train a smaller ViQ-0.4B variant based on SigLIP2-SO400M, so that its parameter count is matched to medium-sized continuous encoders. Under the same protocol as Table~1 of the main paper, ViQ-0.4B attains an overall average accuracy of $56.8$, still outperforming continuous baselines with fewer than $1$B parameters. This indicates that the gains of ViQ stem from the text-aligned pre-training and quantization design rather than from a larger backbone alone.

\section{Additional Analysis on Quantization Methods}
To complement the quantization-method comparison in Table~3c of the main paper, we evaluate two additional vector quantization variants under the same proximal-representation setting, pairing each method with the regularization norm best aligned with its formulation ($L_1$ for LFQ~\cite{yulanguage} and $L_2$ for vanilla VQ~\cite{gersho2012vector}). Vanilla VQ reaches an average score of $65.7$, slightly below SimVQ~\cite{zhu2025addressing} ($66.6$), while LFQ reaches $66.8$, above SimVQ but still clearly below FSQ~\cite{mentzerfinite} ($68.7$). These results reinforce our observation that parameter-free quantizers optimize more stably within our framework than codebook-learning methods. We further find that the regularization norm must match the quantization geometry: replacing the $L_2$ norm with an $L_\infty$ norm for SimVQ degrades its score from $66.6$ to $66.1$, whereas the same $L_\infty$ norm is precisely what benefits the grid-structured FSQ. This indicates that tightly aligning the continuous proximal representation space with the explicit quantization boundaries—rather than the learnability of the codebook alone—is the key factor for preserving information through quantization.

\section{Encoder Memory and Throughput}
\begin{figure}[htbp]
\centering
\includegraphics[width=\linewidth]{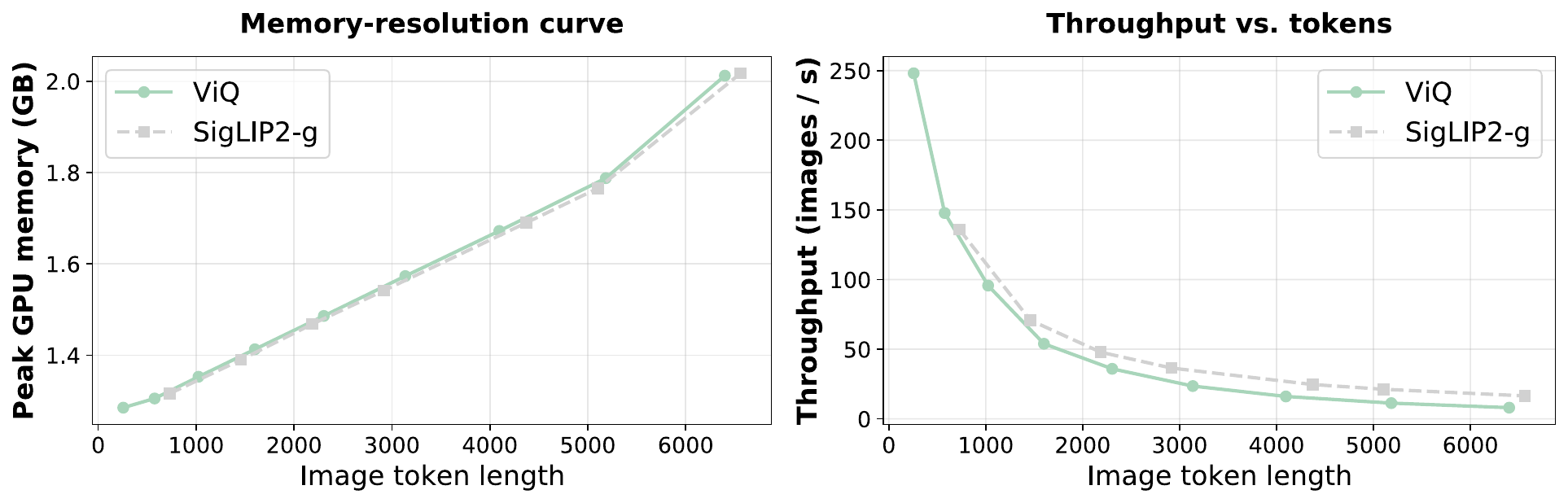}
   \caption{ \small \textbf{Encoder Memory and Throughput.} Peak GPU memory and throughput of ViQ and SigLIP2-g across image token lengths.}
   \label{fig:efficiency}
\end{figure}
We further profile the encoder itself in terms of peak GPU memory and throughput, as shown in Figure~\ref{fig:efficiency}. In memory footprint, ViQ closely matches SigLIP2-g and scales near-linearly with the image token length, indicating that the proximal bottleneck and quantization introduce negligible memory overhead. In throughput, ViQ is on average about $15\%$ slower than SigLIP2-g, a modest cost mainly introduced by the any-resolution processing, the 2D RoPE computation, and the quantization bottleneck. Notably, this overhead is incurred only once during offline code extraction and is fully amortized at training time, where loading the precomputed codes yields the large speed-ups reported in the main paper.

\begin{figure}[htbp]
\centering
\includegraphics[width=\linewidth]{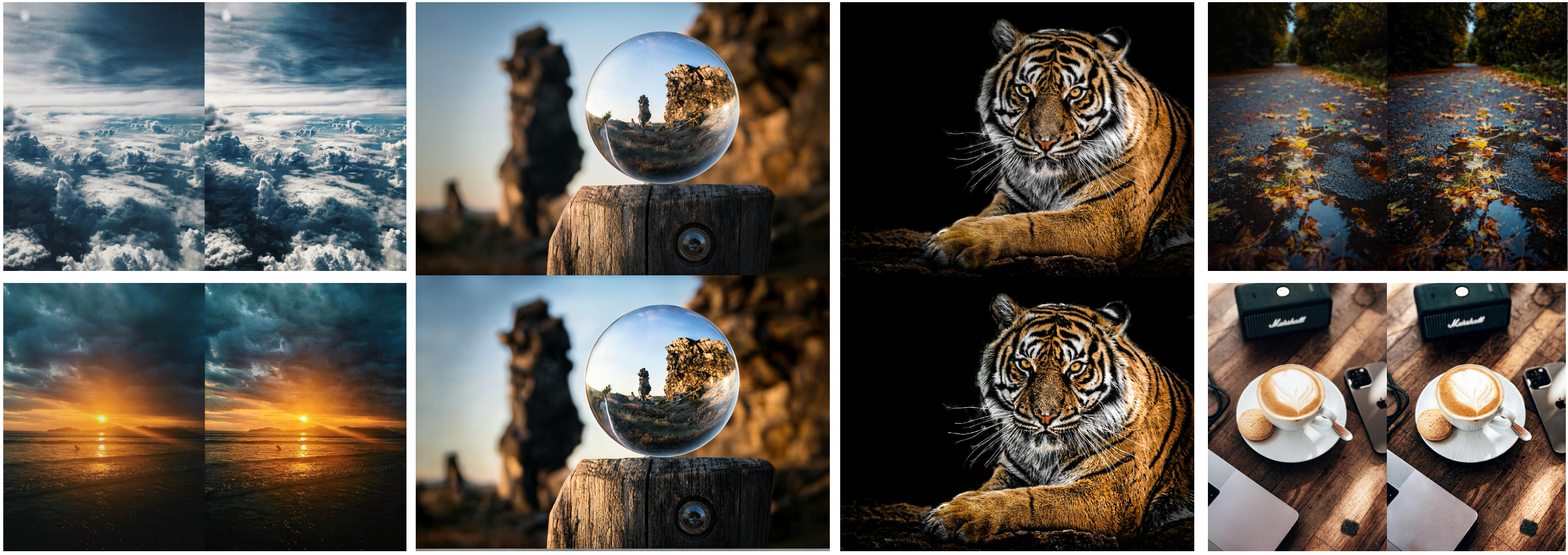}\vspace{-5pt}
    \caption{\textbf{More Reconstructed Visualization Samples of ViQ at Any Resolution. } Left or above is the original image.}
    \label{fig:supp_vis}
\end{figure}

\section{More Visualizations}
We selected images with different resolutions and themes, containing a lot of details, to showcase the reconstruction effects with our ViQ.

\end{document}